\documentclass{article} 
\usepackage{tccml_iclr2025_conference,times}

\usepackage{amsmath,amsfonts,bm}

\def\eqref#1{equation~\ref{#1}}

\def\1{\bm{1}}

\DeclareMathAlphabet{\mathsfit}{\encodingdefault}{\sfdefault}{m}{sl}
\SetMathAlphabet{\mathsfit}{bold}{\encodingdefault}{\sfdefault}{bx}{n}

\usepackage{graphicx}
\usepackage{hyperref}
\usepackage{url}
\usepackage{xcolor}
\usepackage{amsmath}
\usepackage{wrapfig}
\usepackage{subcaption}
\DeclareCaptionLabelFormat{withfigure}{(\arabic{figure}#2)}

\title{Heterogeneous graph neural networks for species distribution modeling}

\author{\bf 
Lauren Harrell$^a$$^1$,
Christine Kaeser-Chen$^b$,\\ \bf 
Burcu Karagol Ayan$^b$,
Keith Anderson$^b$,
Michelangelo Conserva$^a$, \\ \bf 
Elise Kleeman$^a$, 
Maxim Neumann$^b$,
Matt Overlan$^b$,
Melissa Chapman$^a$, \\ \bf 
Drew Purves$^b$ \\ \\
$^a$Google Research
$^b$Google DeepMind \\
$^1$Corresponding author: laurenharrell@google.com
}

\iclrfinalcopy 
\begin{document}

\maketitle

\begin{abstract}
Species distribution models (SDMs) are necessary for measuring and predicting occurrences and habitat suitability of species and their relationship with environmental factors. We introduce a novel presence-only SDM with graph neural networks (GNN). In our model, species and locations are treated as two distinct node sets, and the learning task is predicting detection records as the edges that connect locations to species. Using GNN for SDM allows us to model fine-grained interactions between species and the environment. We evaluate the potential of this methodology on the six-region dataset compiled by National Center for Ecological Analysis and Synthesis (NCEAS) for benchmarking SDMs. For each of the regions, the heterogeneous GNN model is comparable to or outperforms previously-benchmarked single-species SDMs as well as a feed-forward neural network baseline model. 
\end{abstract}

\section{Introduction}
Understanding where species are distributed across the globe is essential not only for advancing our scientific understanding of ecological processes \citep{guisan2011sesam}, but also to enable the strategic implementation of conservation measures and policies \citep{guisan2013predicting}. Species Distribution Models (SDMs) provide a means for predicting species occurrence across landscape scales, generally leveraging species occurrence data and environmental covariates. While a multitude of methods have been developed and explored to predict species distributions \citep{elith2009species, johnston2020estimating, anderson2012harnessing}, many challenges persist. First, data on species is often presence-only (only provides information about where a species is found, not where it is absent), making it difficult to identify the environmental conditions that define the boundaries of a species range. Second, underlying biodiversity data used to train SDMs is often biased \citep{chapman2024biodiversity, Hughes2021-wn}, risking that models reflect these data disparities or observational processes, rather than underlying ecological processes intended \citep{a2022species}. Third, SDMs are difficult to evaluate at scale, due to the lack of ``ground-truth" data \citep{Beery2021-kk}.

A number of recent works have focused on addressing the challenge of utilizing presence-only records for SDMs \citep{Cole2023-zw, Valavi2022-pr}. In this paper, we focus on learning the interaction between the environment and species, the first step towards understanding ecological processes. We present a novel approach to SDMs by using heterogeneous graph neural networks (GNN). We model locations and species as graph nodes, and their interactions as explicit graph edges in our GNN. This enables us to learn the representations of location, species, and their interactions, supervised by the species occurrences records. The GNN model is extensible to different observation and species data modalities and can model various ecological interactions as distinct edge types. Our proof-of-concept results on a benchmark dataset are comparable with common SDM approaches as well as an alternative deep-learning approach.

\section{The case for Graph Neural Networks for Species Distributions}
Graph Neural Networks (GNNs) are a class of deep learning models that capture relationships between entities, by processing data structured as graphs, where nodes represent entities, and edges represent their connections \citep{Scarselli}. Broadly speaking, a graph $\mathcal{G}$ is defined as a set of nodes and edges, $\mathcal{G} = (\mathcal{V},\mathcal{E})$, where $\mathcal{V}$ represents the set of nodes and $\mathcal{E}$ the set of edges. Edges are directed when there is an ordering or dependency that is not symmetric between two nodes and undirected when the relationship between two nodes is the same in both directions. A key feature of GNNs is the ability to pass messages between nodes through the edge sets, resulting in node representations that aggregate information from their neighboring nodes. 

Heterogeneous graphs can contain multiple types of nodes and edges, as opposed to a single node and edge type in homogeneous graphs. Representing data through heterogeneous graphs unlocks flexibility in data structures and ability to learn through multiple types of relationships. Recent development of deep-learning approaches to SDM problems treat the species detection/non-detection or presence/absence as a label set \citep{Cole2023-zw, Estopinan2022-ia, gillespie2024deep}, which limits the information that may be relevant from species-specific features such as taxonomic hierarchy, size, mass, or trophic relationships. Heterogeneous graphs, on the other hand, can represent a variety of data without requiring all observations being represented as feature vectors with the same length, allowing the flexible incorporation of information into the model such as species traits across different taxonomic groups and complex remote-sensing feature data for locations. We can also represent and learn from multiple types of relationships, such as taxonomic similarity and location proximity, to better represent the ecological systems.

\begin{figure}[ht]
\begin{center}
\includegraphics[width=\linewidth]{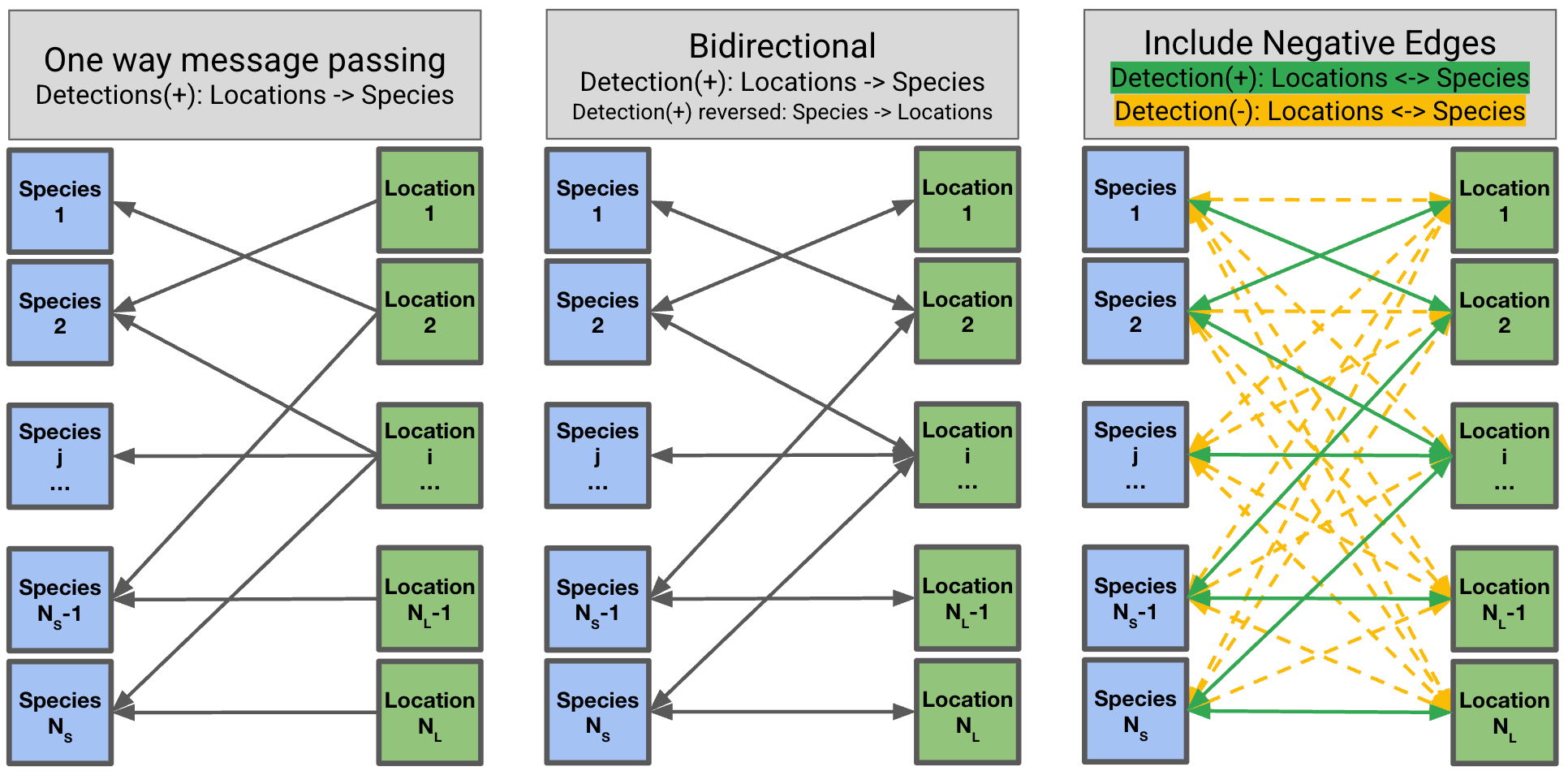}
\end{center}
\caption{Example of bipartite heterogeneous graph structure where species nodes are connected to location nodes through detection edges. The first graph on the left shows one way message passing from locations $v^L_i\in\mathcal{V}^L$ to species $v^S_j\in\mathcal{V}^S$ through detections $\mathcal{E}^{L2S}_{\text{Detection}}$. The middle bipartite graph includes the reversed edge set $\mathcal{E}^{S2L}_{\text{Detection}}$ that sends information from species nodes back to location nodes. The graph on the right includes message passing through (pseudo)-negative edges as a distinct edge set.}
\label{fig:sdm_bipartite}
\end{figure}

We represent the observed species detection records as a heterogeneous graph $\mathcal{G}(\mathcal{V}^S,\mathcal{V}^L,\mathcal{E}^{L2S}_{\text{Detection}}, \mathcal{E}^{S2L}_{\text{Detection}})$ with distinct node sets for species $\mathcal{V}^S$, and locations $\mathcal{V}^L$. We represent detections as edges ($\mathcal{E}^{L2S}_{\text{Detection}}, \mathcal{E}^{S2L}_{\text{Detection}}$) that connect species to location nodes in both directions. For location $v^L_i\in\mathcal{V}^L$ and species $v^S_j\in\mathcal{V}^S$, we include edges $e^{L2S}_{i, j}$ and $e^{S2L}_{j, i}$ in the edge set if species $j$ is detected at location $i$ as shown in Figure \ref{fig:sdm_bipartite}. Our learning task is a link-prediction task \citep{zhang2018link} to infer whether an unseen "detection" edge $e^{L2S}_{i', j'}$ should exist between location node $v^L_{i'}$ and species node $v^S_{j'}$. One way to compute the probability of $e^{L2S}_{i', j'}$ is to decode from learned representations of $v^L_{i'}$ and $v^S_{j'}$, where the species representation is computed by aggregating information from the location nodes with known detections through $\mathcal{E}^{L2S}_\text{Detection}$. More details can be found in Appendix \ref{encoder_appendix}. The training graph may also include a separate non-detection edge set to enable message passing between species and locations without observed detections, such that separate model weights can be learned for aggregating information along these edges.

\section{Application of GNNs to the NCEAS Benchmarks}
We apply the GNN approach to the benchmarks from \cite{Elith2020-io} (which we will refer to as ``NCEAS benchmarks"). The data is comprised of surveys from six distinct regions\footnote{The regions are: 
(1) Ontario, Canada (CAN), (2) South American countries (SA), (3) Switzerland (SWI), 
(4) New South Wales (NSW), (5) Australian Wet Tropics (AWT), and (6) New Zealand (NZ).}, where presence-only (PO) records are available for training, and presence-absence (PA) records are used for testing. We select this dataset because it has been used to benchmark existing presence-only SDMs \citep{Valavi2022-pr} and is publicly available.

For each of the six regions, the dataset provides between 11 and 13 pre-extracted environmental features, in addition to the spatial coordinate information for each data point. The goal is to predict the presence-absence status of between 20 and 54 species, depending on the region, given the presence-only detections in the training data. The characteristics of the training and test data for each region can be found in the Appendix in Table \ref{tab:nceas}. The dataset also supplies 10,000 sampled background locations per region as part of the training data.

\subsection{Graph Data Representation}
We build a heterogeneous graph with two different node types to represent the data. One node set represents the locations, with provided environmental attributes as features. The other node set represents the species. Features for each species node include the species ID encoded as a one-hot vector, and optionally the group information (e.g., plant or bird), for two of the NCEAS regions (NSW and AWT) where species group is provided (Figure \ref{fig:nceas_data} in Appendix). We create an edge set for detections between each location-species pair where a record exists in the PO data. For locations with multiple detections and potential variation in the environmental features, we aggregate the location features using the mean. Species distribution is then learned through an link-prediction task on the graph for whether a detection edge should exist between a location node and a species node. 

\subsection{Model Training}
We train separate GNN models for each region. During training, we train on the graph built from the training PO data with the background points appended as additional location nodes. Message passing on the known detection edges can be modeled as one-way from location to species, or bi-directional where there is also message passing from species nodes back to location nodes. We report results on both model types below.

Species detection at locations is formulated as a link prediction task on this graph. For each target $<v^L_{i'}, v^S_{j'}>$ node pair, we compute a score based on the dot product of the output latent features for the species and location nodes after message passing steps, and apply a sigmoid function on the score to compute the probability of the target link. 

At each epoch during training, we sample training examples of target links with both positive (i.e., known detections) and negative labels. Like most presence-only SDMs \citep{Cole2023-zw, Zbinden_2024, Barbet-Massin2012-ff}, we construct pseudo-negative data by assuming unobserved species are not present in PO locations and all species are not present in background locations. In our experiment, we can utilize the full set of positive examples from the PO dataset in each epoch and also include sampled pseudo-negative examples. The ratio of pseudo-negative examples to positive examples is controllable by a hyperparameter. Training loss is computed as the cross-entropy between link probability and the binary target link labels.  

Our GNN model is based on the Interaction Network (IN) architecture \citep{battaglia2016interaction}. In addition to the IN message passing networks $\phi_O$ and $\phi_R$, we also add separate node set and edge set encoders and decoders before and after the message passing steps to facilitating feature learning. All neural network components are modeled with multi-layer perceptrons (MLPs). We ablate the hidden layer size, the number of hidden layers for these MLPs, and also the number of message passing steps in our experiments; details can be found in Appendix \ref{experiment_details}. 

\subsection{Evaluation}
We adopt the area under the receiver operating characteristic curve ($\text{AUC}_\text{ROC}$) metric for evaluation, similar to \cite{Valavi2022-pr}. For each region, we report the average $\text{AUC}_\text{ROC}$ across all species. When evaluating the GNN, we add the test location nodes to the training graph, but do not add any edges connecting them to species, as the detection status is unknown. We run the trained GNN with message passing on this test graph. Without any edges, the test location nodes' latent features are essentially computed by the learned location encoder, while the species nodes still get message passing from the training graph. We then compute the dot product on features between test location nodes and each of the species nodes, and apply the sigmoid function to compute the link probabilities. 

\subsection{Baseline method}
To study the effects of applying the graph structure to SDM, we compare with a feed-forward MLP model as the baseline. The MLP model uses the environmental features from the NCEAS dataset as input, and has a multi-class classification head as the output. To train the model, we apply the same pseudo-negative generation strategy as above: all observed species classes at a given location are assigned positive labels, while all other species are assigned negative labels. We also assign all species negative labels for background locations. The training batches are composed by the presence-only locations and background locations, where the mixing ratio is controlled by a hyperparameter. In our implementation, we normalize all input features to range [-1, 1] and apply a small random noise to each input feature as perturbation during training. The model is trained with the sigmoid cross entropy loss on the output logits. 

We ablate the hidden layer size, the number of hidden layers, and the ratio of sampled negative data per training batch. In the following, we report results on models with hidden layer size of 32 and 4 layers, except for the NSW region in the NCEAS benchmark where a 6-layer MLP performs better. We observe that, contrary to previous findings, mixing pseudo-negative examples from \textit{background} locations (where we assume all species are absent) in training has no benefit to model performance, as long as the \textit{PO} locations have pseudo-negative labels assigned for unobserved species. We plan to study the effect of background sampling more closely in a future work. 

\begin{figure}[ht]
\begin{center}
\includegraphics[width=\linewidth,scale=0.18]{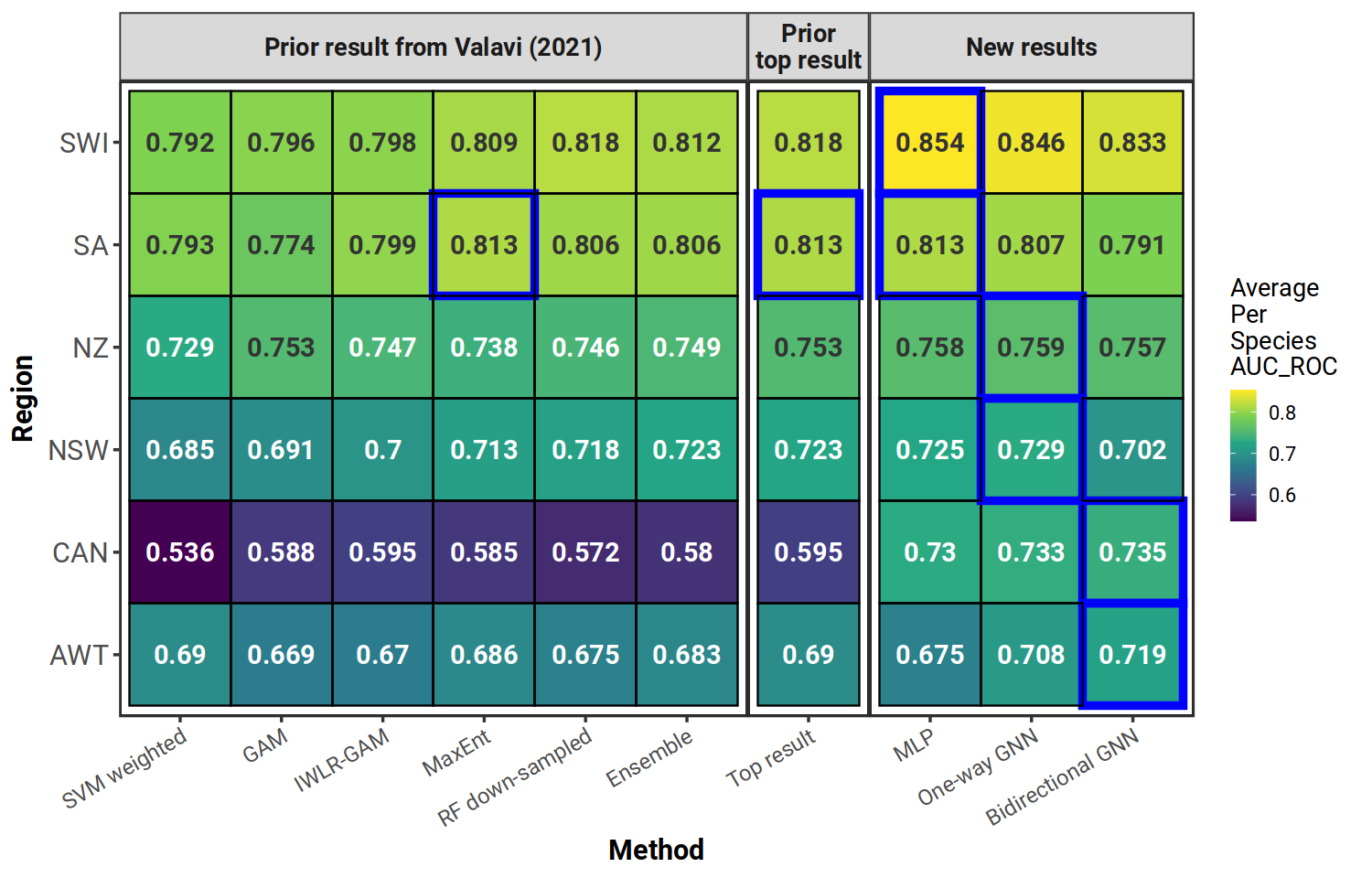}
\end{center}
\caption{$\text{AUC}_\text{ROC}$ averaged across species per site by region and model methodology. The values of prior results were taken from the top scoring models in \cite{Valavi2022-pr}. Top result per region highlighted in blue.}
\label{fig:overall_results}
\end{figure}

\subsection{Results}
A summary of the top $\text{AUC}_\text{ROC}$ models by model type, including GNNs, the baseline MLP, and prior results from \cite{Valavi2022-pr} are displayed in Figure \ref{fig:overall_results}. Some key observations are:

\begin{enumerate}
\item Compared to prior results, GNN has better performance on 3 regions (SWI, CAN, AWT) and comparable performance on 2 regions (NZ, NSW), with a relative increase of 23.5\% over prior methods in CAN. Only on the SA region does the GNN have lower performance. We suspect this is due to the SA dataset being particularly sensitive to over-fitting, which is also observed when training the baseline MLP on this region, in addition to the relatively low number of unique locations in the test data (152 total). 
\item Compared to baseline MLP models, GNN is significantly better on AWT, comparable on 3 regions (NZ, NSW, CAN), and slightly worse on 2 regions (SWI, SA). The performance gain on AWT is particularly promising, as this region does have two taxonomic groups and thus species features beyond the species ID; however the additional species groups in NSW did not produce substantial performance differences.
\item Among the GNN models, we observe variation across regions in the performance trade-off between one-way GNN and bidirectional GNN message passing. Better understanding on the effects of edge directions and graph complexity will be explored in future work. 
\item As a caveat, prior results in \cite{Valavi2022-pr} are all trained as single-species models, and thus do not leverage the potential information shared between species. Both the GNN methods and our baseline MLP method are trained as multi-species models, which likely contributes partly to the observed performance improvements. 
\end{enumerate}

\section{Discussion and future work}
In this work, we provide a basic example of using heterogeneous GNNs for species distribution modeling through a bipartite graph as a proof-of-concept, and show that performance is on par with or exceeds other SDM methods for presence-only data. We are continuing to develop the approach further for more complex, feature-rich data that includes blending presence-only with presence-absence data for model training and graph structures, more informative species traits, and remote-sensing-derived environmental features.  We are also developing improved graph-based learning techniques such as including message-passing through additional edges from species-to-species and location-to-location and weighted loss functions for accounting for different pseudo-negative strategies.

\section{Acknowledgements}
We would like to thank a number of colleagues at Google and Google DeepMind who provided input and support for this work at various stages, including Alvaro Sanchez, Peter Battaglia, Charlotte Stanton, Ben Gaiarin, Michal Kazmierski, Dominic Masters, Tim Coleman, Dan Morris, Jane Labanowski, Dessie Petrova, Chris Brown, Philip Nelson, Tom Denton, Katherine Chou, Mili Sanwalka, Jamie McPike, Grace Joseph, Shelagh McLellan, Grace Coleman, Hugo Larochelle, Kevin Murphy, Sophia Alj, and Maria Abi Raad. We also deeply appreciate our external colleagues who have shared their insights with us, including Neil Burgess, Osgur McDermott Long, Calum Maney, Daniele Baisero, Simon Tarr, and Sara Beery.

\bibliography{references.bib}
\bibliographystyle{iclr2025_conference}

\appendix
\section{Appendix}

\subsection{Software}
Our models were trained using a modified version of the jraph \citep{jraph2020github} software package that was designed for training homogeneous graph neural networks with JAX \citep{jax2018github}, an open-source Python library. An extension of jraph called TypedGraphs was developed for heterogeneous graphs, and deep learning on these graphs was implemented and open-sourced via GraphCast, which developed a GNN for global weather forecasting \citep{lam2023learning, lam2022graphcast}.

\subsection{Model and training methodology details}

Most SDM methodologies that leverage presence-only data are ultimately learning about the location feature distribution of the locations where the species has been observed. For example, the output of the popular SDM method MaxEnt \citep{phillips2006maxent} is effectively a ratio of the joint distribution of the environmental features from presence-only locations to the joint distribution of the background features \citep{elith2011statistical}. Message passing in the GNN is a different way to learn representations of both the species and the locations where, depending on the direction of message passing, species node sets are embedded with aggregating information from the locations and vice versa. For detection edges that go from from locations to species ($\mathcal{E}^{L2S}_{\text{Detection}}$), the species embeddings are updated by aggregating information from locations in which there are detections of that species. If message passing occurs in the reverse direction in $\mathcal{E}^{S2L}_{\text{Detection}}$, the location embeddings are then updated with information aggregated from the species set.

\subsubsection{GNN model steps} \label{encoder_appendix}
Our GNN approach follows the \textit{encode, process, decode} framework established by \cite{Battaglia2018-eh}. Underneath the hood, the encoder models we use in this paper are a multi-layer perceptrons that embed the species node set, location node set, and detection edge set in the same dimensional latent space $\Omega^d$, which is a vector of length $d$ for each node or edge. After encoding, we run a fixed number of message passing steps so that feature information is aggregated and shared through the edges from sender to receiver nodes, following the Interaction Network architecture described in \cite{battaglia2016interaction} and \cite{lam2022graphcast}. 

Our model operates on a heterogeneous graph $\mathcal{G}(\mathcal{V}^S,\mathcal{V}^L,\mathcal{E}^{L2S}_{\text{Detection}})$, which can be expanded to include additional edge sets for message passing to include $\mathcal{E}^{S2L}_{\text{Detection}}$ for bidirectional message passing through detection edges and/or  $\mathcal{E}^{L2S}_{\text{Non-Detection}}$, $\mathcal{E}^{S2L}_{\text{Non-Detection}}$ for message passing for species/location pairs without detections (i.e., ``negative" edges). $\mathcal{V}^S$ represents the node sets indexing the set of species nodes $v^S_j$ and the associated feature set for each species $j$. $\mathcal{V}^L$ is the node set that indexes locations, with a node $v^L_i$ for each location $i$ containing the set of  environmental features. $\mathcal{E}^{L2S}_{\text{Detection}}$ is the edge set that connects location nodes $\mathcal{V}^L$ to species nodes $\mathcal{V}^S$ where there exists detections, and $\mathcal{E}^{S2L}_{\text{Detection}}$ is the edge set that reverses the direction of the detection edges to share information from species $\mathcal{V}^S$ to locations $\mathcal{V}^L$. If included in the model specification, we may also include "negative" detection edges as separate edge sets to aggregate information from location/species pairs without detections $\mathcal{E}^{L2S}_{\text{Non-Detection}}$ and the corresponding reverse direction for bidirectional message passing $\mathcal{E}^{S2L}_{\text{Non-Detection}}$.

\textbf{Step 1: Encode features}. The first step in the GNN embeds the features of all nodes and edges into the same-dimensional latent space. 

\begin{align*}
    \textbf{v}^S_j &= \text{MLP}^{\text{embedder}}_{\mathcal{V}^S}(\textbf{v}^{S,\text{features}}_j) \\
    \textbf{v}^L_i &= \text{MLP}^{\text{embedder}}_{\mathcal{V}^L}(\textbf{v}^{L,\text{features}}_i) \\
    \textbf{e}^{L2S,\text{Detection}}_{\mathcal{V}^L \rightarrow \mathcal{V}^S} &= \text{MLP}^{\text{embedder}}_{\mathcal{E}^{L2S}_{\text{Detection}}}(\textbf{e}^{L2S,\text{Detection},\text{features}}_{\mathcal{V}^L \rightarrow \mathcal{V}^S}) \\
    \textbf{e}^{S2L,\text{Detection}}_{\mathcal{V}^S \rightarrow \mathcal{V}^L} &= \text{MLP}^{\text{embedder}}_{\mathcal{E}^{S2L}_{\text{Detection}}}(\textbf{e}^{S2L,\text{Detection},\text{features}}_{\mathcal{V}^S \rightarrow \mathcal{V}^L}) \quad (optional) \\
     \textbf{e}^{L2S,\text{Non-Detection}}_{\mathcal{V}^L \rightarrow \mathcal{V}^S} &= \text{MLP}^{\text{embedder}}_{\mathcal{E}^{L2S}_{\text{Non-Detection}}}(\textbf{e}^{L2S,\text{Non-Detection},\text{features}}_{\mathcal{V}^L \rightarrow \mathcal{V}^S}) \quad (optional) \\
    \textbf{e}^{S2L,\text{Non-Detection}}_{\mathcal{V}^S \rightarrow \mathcal{V}^L} &= \text{MLP}^{\text{embedder}}_{\mathcal{E}^{S2L}_{\text{Non-Detection}}}(\textbf{e}^{S2L,\text{Non-Detection},\text{features}}_{\mathcal{V}^S \rightarrow \mathcal{V}^L}) \quad (optional)
\end{align*}

\textbf{Step 2: Processing/message passing.} In the next step, our model conducts message passing and node/edge embedding updates using the aggregated features.
For each step in the number of message passing steps, the node and edge latent feature sets are updated using the interaction network, starting with updating the edge sets based on the information of the sender/receiver node pairs. For example, for the detection edge set from locations to species  ($\mathcal{E}^{L2S}_{\text{Detection}}$) included in all model versions, the update is computed as follows: 

\begin{align*}
\textbf{e}^{L2S,\text{Detection}'}_{\mathcal{V}^L_i \rightarrow \mathcal{V}^S_j} &= \text{MLP}^{\text{processor}}_{\mathcal{E}^{L2S}_{\text{Detection}}}([\textbf{e}^{L2S,\text{Detection}}_{v^L_i \rightarrow v^S_j}, {\textbf{v}^L_i, \textbf{v}^S_j}])
\end{align*}

Note that negative edges would have similarly structured but separately learned updates, which allows for different model weights to apply to the aggregated information from the negative edges versus the positive edges. After the edge sets updates are computed, the receiver node sets (in this case the species node set) is updated by aggregating the information across the edges arriving at that node. Note that for aggregation of the edge information, if $w_k = 1$, this corresponds to the segment sum, and if $w_k = \dfrac{1}{n_e}$ we are taking the segment mean. 

\begin{align*}
{\textbf{v}^{S'}_j} &= \text{MLP}^{\text{processor}}_{\mathcal{V}^S}(\textbf{v}^S_j, \sum_{\textit{e}^{L2S,\text{Detection}}_{v^L_i \rightarrow v^{S'}_j}} w_k \textbf{e}^{L2S,\text{Detection}}_{v^L_i \rightarrow v^{S'}_j})
\end{align*}

If message passing is bidirectional, then the location nodes are updated in a similar fashion as the species nodes. However, in the one-way graph, the location features do not receive messages from the species nodes and are updated independently:

\begin{align*}
    {\textbf{v}^L_i}' &= \text{MLP}^{\text{processor}}_{\mathcal{V}^L}(\textbf{v}^{L}_i) \\
\end{align*}

The new feature values are updated with a residual connection. 
\begin{align*}
\textbf{e}^{L2S,\text{Detection}}_{\mathcal{V}^L_i \rightarrow \mathcal{V}^S_j} &\leftarrow \textbf{e}^{L2S,\text{Detection}}_{\mathcal{V}^L_i \rightarrow \mathcal{V}^S_j} + \textbf{e}^{L2S,\text{Detection}}_{\mathcal{V}^L_i \rightarrow \mathcal{V}^{S'}_j} \\
\textbf{v}^S_j &\leftarrow \textbf{v}^S_j + {\textbf{v}^{S'}_j} \\
\textbf{v}^L_i &\leftarrow \textbf{v}^L_i + {\textbf{v}^{L'}_i}
\end{align*}

\textbf{Step 3: Decoding and output.} The resulting output embeddings for the species and location nodes are encoded into a fixed size of the same dimension (defined in Table \ref{table:parameter_sweep_config}). In order to learn a binary classifier for whether an edge between location $i$ and species $j$ should exist, we need to define a function to decode to a scalar for binary classification $f(\textbf{v}^L_i, \textbf{v}^S_j) \to z_{L_i,S_j}$. For the purposes of this work, we use a simple dot product decoder as the similarity between the latent vectors between the species embeddings and location embeddings to decode create a single scalar to represent the score for the potential edge where $z_{L_i,S_j} = \textbf{v}^L_i\cdot\textbf{v}^S_j$. In the future, we could use the MLP embeddings of the edges more directly or learn alternative scoring functions that reflect the divergence of the positive versus negative edge embeddings. 

\subsubsection{Future architectures} Presented in this paper is a simple proof-of-concept of leveraging heterogeneous graph neural networks where message passing occurs only on the bipartite graph between species and location node sets. In future work, we are expanding on this methodology for heterogeneous graph learning where we can separate out the encoding and processing steps of the GNN more explicitly where initial encoding occurs along the bipartite species $\Leftrightarrow$ locations subgraph, but additional processing message passing models can be applied through location $\Leftrightarrow$ location and/or species $\Leftrightarrow$ species edge sets that can enable smoothing and improved representations to address sampling bias for locations or species without detections in the data.

\subsubsection{Loss function: Binary cross entropy loss with logits} 
We start with the sigmoid binary cross entropy loss with logits, which is a common loss function for link prediction tasks. For the $k$th edge representing species $j$ and location ${i}$, let $score_{k}$ be the model output and $y_{k}$ represent the true label of 1 for a detection present in the data and 0 otherwise, then for edge $k$, the loss is computed as:

\begin{align*}
   prob_k &= \text{sigmoid}(score_k) \\
   loss_k &= y_k*\text{log}(prob_k) + (1-y_k)*\text{log}(1 - prob_k)
\end{align*}

This loss function increases when scores are diverging from their expected label (i.e., high scores for "negatives" and low scores for positives). We can aggregate this loss function over all the edges using either the mean or the sum; in our examples we use the mean. 

Currently our loss function does not differentiate between different pseudo-negative types such as the loss functions proposed in similar deep learning \citep{Cole2023-zw, Zbinden_2024}. It is also computed by taking the overall mean rather than averaging across loss at a species level, which may be impacting the performance on per-species $AUC_{ROC}$ versus an overall $AUC_{ROC}$ aggregated across all edges irrespective of species. To optimize for improved performance at the species level and to minimize effects of imbalances in the number of occurrence records per species, future work should experiment with alternative loss functions and consider optimizing per-species rather than across all edges weighted equally. 

\subsection{Hyperparameter ablation experiment settings}\label{experiment_details}
We experimented with a number of potential options for ablation
studies to train the GNN models. 
\begin{table}[ht]
\centering
\begin{tabular}{|| l | c ||} 
 \hline
 Configuration or parameter & Options  \\ [0.5ex] 
 \hline\hline 
 Include negative edges for message passing & [True, False] \\
 Message passing direction &  [One-way, Bidirectional] \\
 Latent sizes & [16, 20, 32, 64]  \\
 Number of hidden layers & [1, 2, 3] \\
 Number of message passing steps & [1, 2, 3] \\
 Negative sampling strategy & [Uniform, Stratified ($K_{Locations}$ per species)] \\ 
 Number of negative edges sampled per training step & [$\#$ of positive edges, 1000] \\
 Proportion of negatives from PO locations (versus background) & [Random, 0.75, 1.0] \\
 Aggregation of edges for node updates & [Segment sum, Segment mean] \\ 
 Learning rates & [0.001, 0.0001] \\
 Activation functions & [relu, leakyrelu, softplus, silu, hardsilu, sparseplus] \\ [1ex] 
 \hline
\end{tabular}
\caption{Hyperparameter configuration options for GNN models}
\label{table:parameter_sweep_config}
\end{table}

\subsection{Additional notes on benchmarking on NCEAS data}
The NCEAS data cover six distinct regions of the world where there exists disparate sets of presence-only records along with high-quality presence-absence data in the following regions:
\begin{itemize}
    \item Ontario, Canada (CAN),
    \item South American countries (SA),
    \item Switzerland (SWI),
    \item New South Wales (NSW),
    \item Australian Wet Tropics (AWT), and
    \item New Zealand (NZ).
\end{itemize}

The number of species, number of location variables, unique locations with detections in the training data (excluding the additional 10,000 background locations), and unique locations in the presence-absence test data are shown in Table \ref{tab:nceas}. 

\begin{table}[ht]
    \centering
    \begin{tabular}{||p{0.07\linewidth}  | p{0.30\linewidth} |p{0.30\linewidth} |p{0.11\linewidth}  |p{0.11\linewidth}  | p{0.09\linewidth}  ||}
\hline
Region Code	&	Region	&	Species set	&	Number of location variables	&	Training Locations (with detections)	&	Unique Test Locations	\\ \hline \hline
AWT	&	Australian Wet Tropics, Queensland, Australia	&	20 birds; 20 vascular plants	&	13	&	3806	&	442	\\ \hline
CAN	&	Ontario, Canada	&	30 birds	&	11	&	5063	&	14571	\\ \hline
NSW	&	North-east New South Wales, Australia	&	7 bats; 8 diurnal birds; 2 nocturnal birds; 8 open-forest trees; 8 open-forest understorey vascular plants; 7 rainforest trees; 6 rainforest understory vascular plants; 8 small reptiles	&	13	&	3323	&	8746	\\ \hline
NZ	&	New Zealand	&	52 vascular plants	&	13	&	3088	&	19120	\\ \hline
SA	&	Continental Brazil, Ecuador, Colombia, Bolivia, and Peru, South America	&	30 vascular plants	&	11	&	1178	&	152	\\ \hline
SWI	&	Switzerland	&	30 trees	&	13	&	11429	&	10013	\\
\hline
    \end{tabular}
    \caption{Characteristics of the NCEAS datasets for each region}
    \label{tab:nceas}
\end{table}

\begin{figure}[ht]
\begin{center}
\includegraphics[width=1.0\textwidth,scale=0.5]{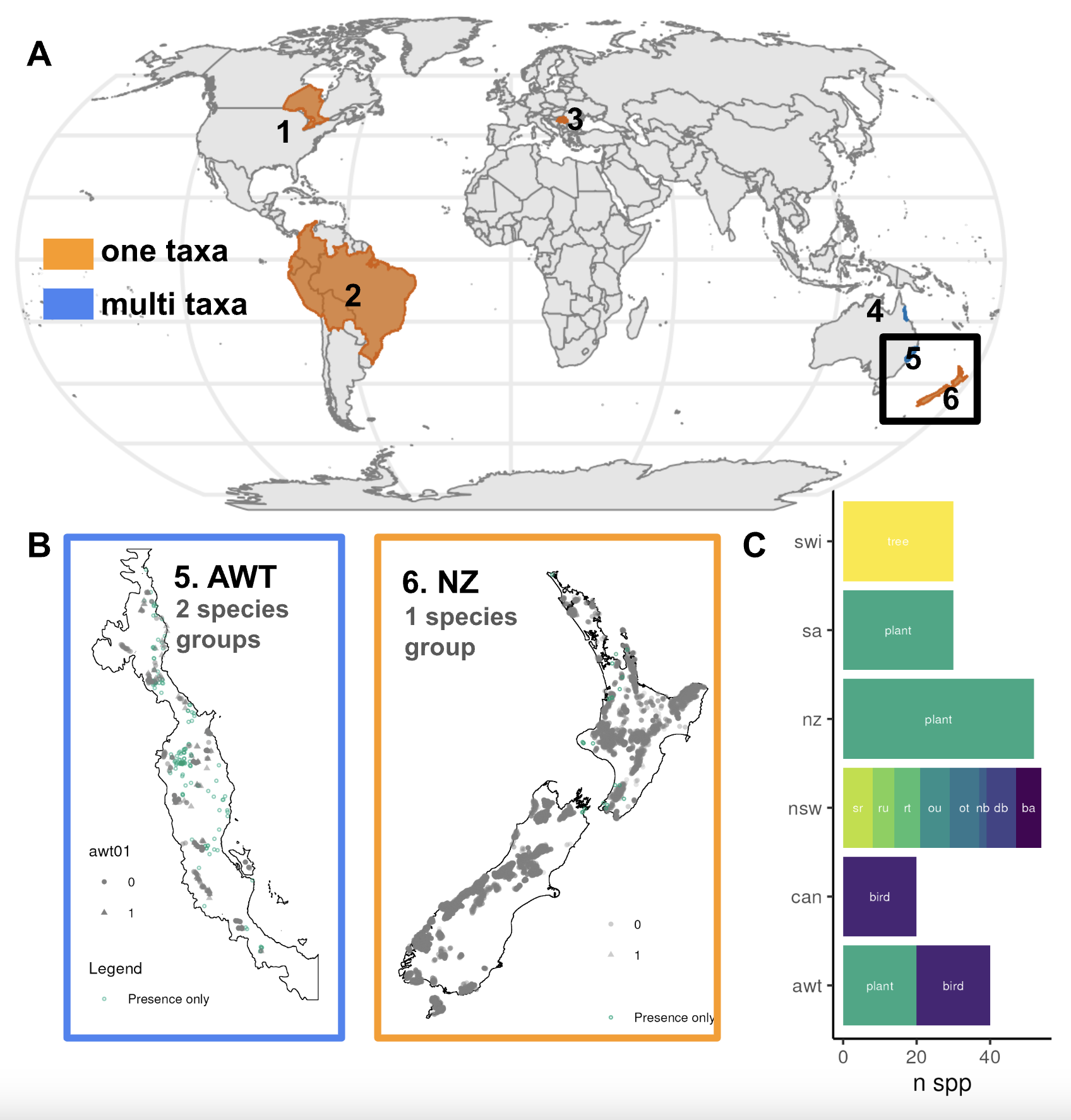}
\end{center}
\caption{NCEAS Data. (A) The locations of each of the six regions around the globe shaded by number of taxonomic groups. (B) Example presence and presence/absence observations for one species in each of the SWT and NZ regions. (C) Counts of species per taxonomic group in each region.}
\label{fig:nceas_data}
\end{figure}

\subsubsection{Limitations of NCEAS data}
The NCEAS data provide a common set of species detection records for which different statistical and machine learning algorithms can be compared directly. Given that the majority of species distribution models are designed for use with presence-only data available for training, it is valuable to have this data that includes examples of presence-only training data alongside a held-out set of presence-absence data that can be used to directly examine how the model would perform against explicit non-detection records. With that said, it has a few important limitations:
\begin{enumerate}
    \item Species detections were made over potentially long time frames, with no recent records after 2006. Some detections in the data occurred over 100 years ago. 
    \item Environmental covariates for the locations are fixed (not time-dynamic) and may not reflect the conditions at the time of detection. Alternatively, the measurements of a given detection record may not be comparable to measurements taken at other locations as counter-factuals if measured at different times. 
    \item Models trained/evaluated on this data assume the relationship between the species and location (habitat suitability process) remains constant over time (i.e., that species don't change their habitat niche needs and their suitable habitat would remain the same). 
    \item The environmental variables measured and collected may not be sufficient for predicting species presence. In practice, there will be an upper bound limit to model predictive performance if key variables are missing from the data.
    \item No species-specific information - any identifying information about the species is obscured, and thus little or no species traits can be included as features. 
    \item Species set contained are from common taxonomic groups that are more likely to be sampled in citizen science (e.g., birds and trees) and from regions that are more likely to be represented more generally in large-scale databases like the Global Biodiversity Information Facility \citep{GBIF_general}. Models that perform well for these species and regions may not be the best performing models for rare species or under-characterized regions. 
\end{enumerate}
\subsection{Limitations of benchmarking comparisons}
The benchmarking study on the NCEAS data was limited to only methods designed for single species models \citep{Valavi2022-pr}. Direct comparison of results of our GNN, which is trained on all available species, to models trained one species at a time have slightly different objectives and are leveraging information in different ways, and thus performance differences may not be isolated to the methodology alone but how the features and training data are processed. We are also not comparing with many other deep-learning methodologies that simultaneously estimate species distributions across multiple species, and thus we cannot show definitively that the GNN approach outperforms an alternative neural network approach. With that said, by showing our results in relation to existing benchmarking results from the single species models along with a baseline MLP model, we demonstrate the promise of this model for future applications. 

Additionally, evaluating on presence-absence data is not the same as true species occupancy/non-occupancy of a given location. Absence data doesn't mean a particular species doesn't occupy a given location, just that it wasn't observed there during the survey or set of surveys included. Presence-absence data with repeated visits can be used to construct alternative types of models such as Bayesian occupancy models that aim to separate the detection process from the habitat niche process. Using presence-absence data as a benchmarking or outcome target may still create distortions or biases in the resulting inferences, particularly for species that are much more difficult to detect even when they do occupy a given location.


\end{document}